\pdfoutput=1

\documentclass[11pt]{article}

\usepackage[final]{acl}

\usepackage{microtype}
\usepackage{hyperref}
\usepackage{url}
\usepackage{booktabs}
\usepackage{times}
\usepackage{latexsym}
\usepackage{amsmath}
\usepackage{amsfonts}
\usepackage{graphicx}
\usepackage{multirow}
\usepackage{subcaption}
\usepackage[T1]{fontenc}

\usepackage{tikz}

\usepackage{listings}
\usepackage{color}
\definecolor{dkgreen}{rgb}{0,0.6,0}
\definecolor{gray}{rgb}{0.5,0.5,0.5}
\definecolor{mauve}{rgb}{0.58,0,0.82}
\definecolor{lightblue}{rgb}{0.9, 0.95, 1.0}
\definecolor{lightgray}{rgb}{0.9, 0.9, 0.9}

\usepackage{colortbl}

\usepackage{algorithmic}
\usepackage{algorithm}

\usepackage[utf8]{inputenc}
\usepackage{multirow}
\usepackage{microtype}
\usepackage{booktabs}

\title{Pruning General Large Language Models into Customized Expert Models}

\author{
  Yiran Zhao$^{1}$ \; Guizhen Chen$^{2,3}$  \; Kenji Kawaguchi$^{1}$
   \; Lidong Bing$^{4}$ \;  Wenxuan Zhang$^{5}$\footnotemark[2] \\
  $^1$ National University of Singapore \quad  $^2$ Nanyang Technological University, Singapore \\ 
  $^3$ DAMO Academy, Alibaba Group, Singapore \quad $^4$ MiroMind \\
  $^5$  Singapore University of Technology and Design \\
}

\begin{document}

\newcommand{\yiran}[1]{\textcolor{brown}{\it\small [#1 --Yiran]}}
\newcommand{\isaac}[1]{\textcolor{orange}{\textbf{[Isaac:} #1]}}

\maketitle
\renewcommand{\thefootnote}{\fnsymbol{footnote}}
\footnotetext[2]{Wenxuan Zhang is the corresponding author.}
\renewcommand{\thefootnote}{\arabic{footnote}}

\begin{abstract}

Large language models (LLMs) have revolutionized natural language processing, yet their substantial model sizes often require substantial computational resources. To preserve computing resources and accelerate inference speed, it is crucial to prune redundant parameters, especially for experienced users who often need compact expert models tailored to specific downstream scenarios. However, most existing pruning methods focus on preserving the model’s general capabilities, often requiring extensive post-training or suffering from degraded performance due to coarse-grained pruning. In this work, we design a \underline{Cus}tom \underline{Prun}ing method (\texttt{Cus-Prun}) to prune a large general model into a smaller lightweight expert model, which is positioned along the ``language'', ``domain'' and ``task'' dimensions. By identifying and pruning irrelevant neurons of each dimension, \texttt{Cus-Prun} creates expert models without any post-training. 
Our experiments demonstrate that \texttt{Cus-Prun} consistently outperforms other methods, achieving minimal loss in both expert and general capabilities across various models from different model families and sizes.\footnote{Our code is publicly available at \url{https://github.com/zhaoyiran924/Custom-Prune}.}

\end{abstract}

\section{Introduction}

Large language models (LLMs)~\citep{achiam2023gpt, reid2024gemini, dubey2024llama, team2024gemma} have revolutionized the field of natural language processing (NLP), emerging as powerful tools with widespread applications across various languages~\citep{cui2023efficient, yang2024qwen2, seallms3}, domains \citep{ li2023huatuo26m, roziere2023code, li2023large}, and tasks~\citep{azerbayevllemma, alves2024tower,naacl-sentiment}.
However, the impressive performance of LLMs often comes at the cost of immense model sizes, mostly containing billions of parameters and thus demand significant computing resources~\citep{goldstein2023generative, musser2023cost}. To address this issue, researchers have recently proposed various model pruning methods for LLMs. These methods aim to reduce model parameters while maintaining the overall performance through techniques such as removal of unimportant structures~\citep{ma2023llm, men2024shortgpt, songsleb}, matrix approximation~\citep{sharmatruth, ashkboosslicegpt}, and extensive post-training after pruning~\citep{wang2024reconstruct, xiasheared}.

Most existing pruning methods focus on preserving the \textit{general capabilities} of the model, often evaluated by broad-spectrum benchmarks like MMLU~\citep{hendrycksmeasuring}.
While aiming for overall versatility, they may not align well with real-world user needs for a pruned small model, which are usually more \textit{specific and targeted}. For instance, a user might require a question-answering model tailored specifically for the education domain in German. 
Such specialized request in fact aligns well with the fundamental motivation behind pruning: to create a smaller model by eliminating unnecessary parameters. In this context, the notion of ``unnecessary'' parameters becomes more precise—referring to those parameters that are irrelevant to the particular use case. By selectively pruning these redundant parameters, one can construct a smaller, expert model that is better aligned with specific requirements.

However, current pruning techniques focusing on preserving the general capabilities of LLMs often employ coarse-grained approaches such as removing entire layers or modules \citep{li2022parameter, kurz2024language, huang2024pruning}. Therefore, it may remove parameters that are critical for specialized downstream scenarios, and thus sometimes require extensive post-training to recover the pruned capabilities \citep{xiasheared, muralidharan2024compact}. On the other hand, the preserved certain capabilities may not be useful or relevant to our desired application scenario. This misalignment with practical requirements, where users need compact ready-to-deploy expert models without retraining, motivates us to design a more fine-grained and expert model targeting approach.

\begin{figure}[t]
    \centering
 \includegraphics[width=0.48\textwidth]{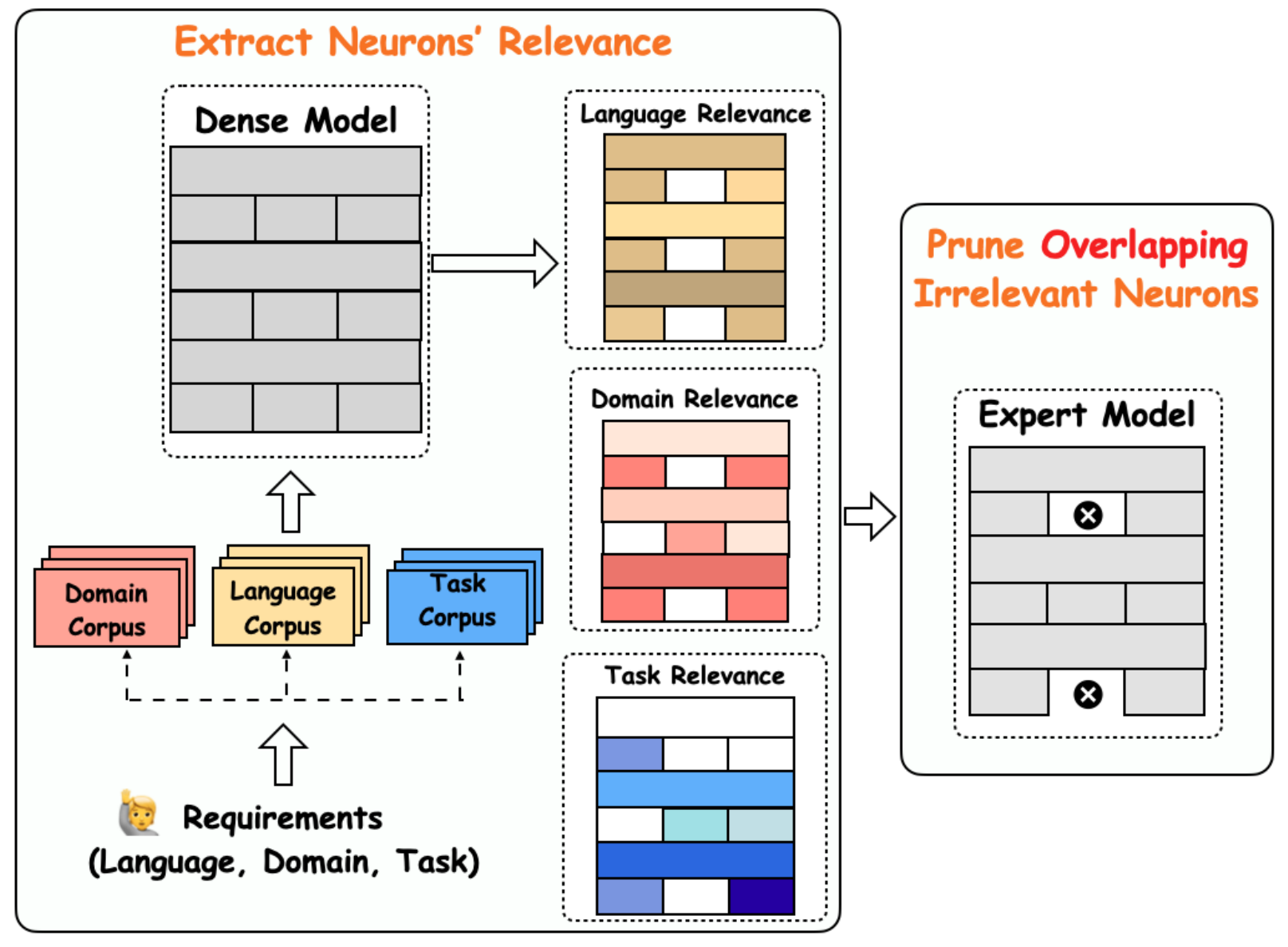}
    \caption{Given a requirement for an expert model across three dimensions (language, domain, and task), \texttt{Cus-Prun} (i) identifies irrelevant neurons for each dimension using corresponding corpora (left), and (ii) prunes \textbf{overlapping} irrelevant neurons across dimensions to obtain the expert model (right).}
    \label{fig:cus-prun}
\end{figure}

In this work, we introduce a novel \underline{Cus}tom \underline{Prun}ing (\texttt{Cus-Prun}) method, designed to prune a large general-purpose model into a small specialized expert model tailored for specific scenarios.
To achieve broad adaptability, we define the expert model by positioning the target user's needs along three key dimensions: language (e.g., English, Chinese, German), domain (e.g., E-commerce, education), and task (e.g., QA, summarization).
Inspired by existing studies that certain neurons are responsible for certain functions \citep{zhao2024large, tang2024language, liang2024multilingual}, \texttt{Cus-Prun} identifies and preserves critical neurons that are more relevant to particular languages, domains, or tasks, while pruning less relevant ones, ultimately leading to a smaller expert models.
Specifically, as illustrated in Figure \ref{fig:cus-prun}, 
\texttt{Cus-Prun} first identifies irrelevant neurons for each dimension by assessing the impact of their removal on the generated output when processing corresponding corpus, which could be easily constructed from the relevant plain text documents.
Next, the expert model is constructed by pruning irrelevant neurons across all dimensions.
Furthermore, \texttt{Cus-Prun}'s flexibility allows it to focus on one, two, or all three dimensions (language, domain, task) as needed, making it adaptable to a wide range of real-world applications where specialized LLMs are required. Importantly, by performing fine-grained pruning at the neuron level, the method could also retain most of the essential neurons within the model backbone, thereby preserving most general capabilities.

We conduct comprehensive experiments to evaluate the performance of \texttt{Cus-Prun} across various scenarios. Experimental results demonstrate that it consistently outperforms other pruning methods in all settings. For three-dimensional specific expert models, \texttt{Cus-Prun} demonstrates remarkable pruning effectiveness across different model families and sizes, such as Mistral-Nemo-12B, Llama3-8B, Llama2-13B, and Llama3-70B, while maintaining strong expert and general capabilities. The method is evaluated across multilingual, multidomain, and multitask datasets, as well as representative compound NLP benchmarks, and consistently outperforms state-of-the-art pruning methods. Moreover, \texttt{Cus-Prun} is highly adaptable, working effectively across a wide range of pruning ratios, even up to nearly half the parameters, without compromising its superior performance. For more focused applications, such as two- or one-dimensional specific expert models (e.g., language-domain specific or language-specific models), \texttt{Cus-Prun} continues to significantly outperform other pruning methods, showcasing its versatility and effectiveness in diverse specialized settings.

\section{Custom Pruning (\texttt{Cus-Prun})}

An expert model could be generally positioned from three dimensions: ``language'' ($L\in\mathbb{L}$), ``domain'' ($D\in\mathbb{D}$), and ``task'' ($T\in\mathbb{T}$), which can be represented as $LLM_{\text{Exp}} := (L, D, T) \in \mathbb{L}\times \mathbb{D}\times \mathbb{T}$. Specifically, the language dimension encompasses various languages such as English, Spanish, and Thai. The domain dimension covers different fields like finance, legal, and medical. The task dimension includes various applications such as question-answering, data-to-text, and summarization. In this section, we propose a custom pruning method named \texttt{Cus-Prun} to derive smaller expert models with flexible customization granularity.

\subsection{Foundational Custom Pruning}

Drawing inspiration from recent LLM interpretation studies~\citep{tang2024language,liang2024multilingual,zhao2024large} that many parameters in the model are redundant to processing a specific ``language'', we hypothesize that this phenomenon can be extended to other dimensions such as ``domain'' and ``task'', meaning that certain parameters remain unused when handling a specific dimension. 
Rather than eliminating entire layers or modules~\citep{songsleb, men2024shortgpt, zhang2024finercut}, \texttt{Cus-Prun} performs a fine-grained investigation by identifying and removing redundant neurons (i.e., individual rows or columns in parameter matrices) across all components (e.g., attention and feed-forward layers).

Concretely, when handling each dimension, we identify a specific set of \textit{irrelevant neurons} in the original LLM, denoted as $\tilde{\mathcal{N}}_{L}$, $\tilde{\mathcal{N}}_{D}$, and $\tilde{\mathcal{N}}_{T}$ for $L$, $D$, and $T$, respectively. 

Specifically, to identify irrelevant neurons corresponding to the selected dimension, we construct a corpus within that dimension while ablating others. For example, to determine irrelevant neurons for a specific language $L_{\text{Exp}}$, we create a corpus set 
\begin{equation}
C_{L_{\text{Exp}}}=\{(L_{\text{Exp}}, D, T)|D\in\mathbb{D}, T\in\mathbb{T}\},
\end{equation} 
comprising documents in language $L_{\text{Exp}}$ across various domains $D$ and tasks $T$. We then identify neurons that are consistently irrelevant across all documents in $C_{L_{\text{Exp}}}$,
\begin{equation}
\scalebox{0.92}{$ 
\tilde{\mathcal{N}}_{L_{\text{Exp}}} = 
\left\{
\text{Neuron} \middle| 
\text{Irrelevant to } c, \; \forall c \in C_{L_{\text{Exp}}}
\right\},
$}
\label{equ:neuron}
\end{equation}
where a neuron is considered irrelevant if its removal from the parameter matrix affects the generated output below a specified threshold. Formally,
for $i$-th neuron in layer $l$, denoted as ${N}_{i}^{(l)}$, its relevance to document $c$ is measured by $|{h}_{\setminus{{N}_{i}^{(l)}}, i}(c) - h_i(c)|_2$, where $h_i(c)$ is the layer output and ${h}_{\setminus{{N}_{i}^{(l)}}, i}(c)$ is the output with the neuron removed. Furthermore, neurons with impact in the lowest $\sigma\%$ are considered irrelevant, where $\sigma$ is a pre-defined pruning ratio.

Similarly, we could establish corresponding corpus sets for other dimensions, 
\begin{equation}
C_{D_{\text{Exp}}}=\{(L, D_{\text{Exp}}, T)|L\in\mathbb{L}, T\in\mathbb{T}\},
\end{equation} 
\begin{equation}
C_{T_{\text{Exp}}}=\{(L, D, T_{\text{Exp}})|L\in\mathbb{L}, D\in\mathbb{D}\},
\end{equation} to extract irrelevant neurons, $\tilde{\mathcal{N}}_{D_{\text{Exp}}} $ and $\tilde{\mathcal{N}}_{T_{\text{Exp}}} $. 
Finally, the expert model could constructed by
\begin{equation}
\scalebox{0.87}{$ 
\mathcal{LLM}_{\text{Exp}} = 
\mathcal{LLM} \ominus 
\left\{
\tilde{\mathcal{N}}_{L_{\text{Exp}}} 
\cap \tilde{\mathcal{N}}_{D_{\text{Exp}}} 
\cap \tilde{\mathcal{N}}_{T_{\text{Exp}}}
\right\},
$}
\label{equ:cus}
\end{equation}
where $\ominus$ represents removing the corresponding neurons from $\mathcal{LLM}$. The overall
algorithm is further illustrated in Algorithm \ref{algo:adaptive_pruning}.

\begin{figure}[t]
\scalebox{0.95}{ 
\begin{minipage}{\linewidth}
\begin{algorithm}[H]
\renewcommand{\algorithmicrequire}{\textbf{Input:}}
\renewcommand{\algorithmicensure}{\textbf{Output:}}
\caption{Adaptive Custom Pruning}
\begin{algorithmic}[1]
\REQUIRE Original language model $\mathcal{LLM}$, request for expert model $\mathcal{LLM}_{\text{Exp}}$ with selected dimensions: $L_{\text{Exp}}$, $D_{\text{Exp}}$, $T_{\text{Exp}}$ (any subset), request for pruning ratio $\sigma$.
\STATE \texttt{// Construct specific corpora for each selected dimension.}
\STATE $C = \{\}$
\IF{$L_{\text{Exp}}$ is specified}
    \STATE $C = C \cup \{(L_{\text{Exp}}, D, T) \mid D \in \mathbb{D}, T \in \mathbb{T}\}$
\ENDIF
\IF{$D_{\text{Exp}}$ is specified}
    \STATE $C = C \cup \{(L, D_{\text{Exp}}, T) \mid L \in \mathbb{L}, T \in \mathbb{T}\}$
\ENDIF
\IF{$T_{\text{Exp}}$ is specified}
    \STATE $C = C \cup \{(L, D, T_{\text{Exp}}) \mid L \in \mathbb{L}, D \in \mathbb{D}\}$
\ENDIF
\STATE \texttt{// Identify irrelevant neurons for each selected dimension.}
\FORALL{neuron $N_{i}^{(l)}$ in $\mathcal{LLM}$}
    \IF{$\forall c \in C,$ $N_{i}^{(l)} \in \tilde{\mathcal{N}}(c)$}
        \STATE $\tilde{\mathcal{N}} \gets \tilde{\mathcal{N}} \cup N_{i}^{(l)}$
    \ENDIF
\ENDFOR
\STATE \texttt{// Prune irrelevant neurons to obtain expert model.}
\STATE $\mathcal{LLM}_{\text{Exp}} = \mathcal{LLM} \ominus \tilde{\mathcal{N}}$
\ENSURE $\mathcal{LLM}_{\text{Exp}}$
\end{algorithmic}
\label{algo:adaptive_pruning}
\end{algorithm}
\end{minipage}
}
\end{figure}

\subsection{Adaptive Custom Pruning}\label{sec:adaptive}

In many applications, the need for expertise might be constrained to one or two dimensions. For example, a language-specific or domain-specific model only requires pruning along a single dimension, while a language-domain-specific model (e.g., a Chinese Medical LLM) constrains two dimensions. In this section, we extend \texttt{Cus-Prun} to support different granularity levels.

\paragraph{Two-Dimensional Specific Expert Model}

Without losing generality, we use the language-domain expert model as a concrete example, which requires an expert model constrained in two dimensions: language ($L_{\text{Exp}}$) and domain ($D_{\text{Exp}}$). We similarly derive the sets of irrelevant neurons $\tilde{\mathcal{N}}_{L_{\text{Exp}}}$ and $\tilde{\mathcal{N}}_{D_{\text{Exp}}}$, and
obtain the expert model by pruning the original dense model as follows: \begin{equation}
\mathcal{LLM}_{\text{Exp}}:=\mathcal{LLM}\ominus\{\tilde{\mathcal{N}}_{L_{\text{Exp}}} \cap \tilde{\mathcal{N}}_{D_{\text{Exp}}}\}.
\end{equation}

\paragraph{One-Dimensional Specific Expert Model}
We use the language-specific expert model as an example, which focuses on optimizing performance for a certain language ($L_{\text{Exp}}$), irrespective of domain or task. Similarly, we obtain the language-specific corpus $C_{L_{\text{Exp}}}$, then identify irrelevant neurons  $\tilde{\mathcal{N}}_{L_{\text{Exp}}}$ 
and extract the expert model by 
\begin{equation}
\mathcal{LLM}_{\text{Exp}}:=\mathcal{LLM}\ominus\{\tilde{\mathcal{N}}_{L_{\text{Exp}}} \}.
\end{equation}
To enhance efficiency, we implement the parallel neuron-detection method~\citep{zhao2024large}, which accelerates the sequential calculations from line14 to line16 in Algorithm \ref{algo:adaptive_pruning}.

\section{Preliminary Evaluation}\label{sec:pre_eva}

In this section, to simulate realistic user-defined requirements where all three dimensions (language, domain, task) are explicitly specified, we conduct preliminary experiments on developing expert models with fine-grained pruning. This setting not only validates our approach under the most demanding conditions but also serves as a prototype for later experiments with one- or two-dimensional customizations.

\paragraph{Experiment Design} 

We consider three scenarios, each named according to the \textit{language}-\textit{domain}-\textit{task} pattern:
\textit{Korean-Legal-Summarization}~\citep{hwang2022multi}, \textit{English-Medical-MCQ}~\citep{garcia2024medical}, and \textit{Chinese-E-commerce-Sentiment Analysis}~\citep{zhang2015character}.
These scenarios were selected because each represents a unique combination of language, domain, and task, effectively simulating the diverse needs a user might specify in practice.

For each scenario, we curate the corresponding corpus for each dimension. This curation can be done through manual collection or by automatically retrieving relevant documents online. In this preliminary study, without loss of generality, we employ a strong proprietary model\footnote{\url{https://platform.openai.com/docs/models/gpt-4o}} to generate a corpus containing $50$ documents for each dimension. Detailed prompts can be found in Appendix \ref{sec:appen_prompt}. The generated documents could then be used to determine the relevance of neurons for each dimension of each scenario.

\paragraph{Experiment Setup}
We use Llama3-8B~\citep{dubey2024llama} as the original dense model and follow the typical setting to set the pruning ratio as $25\%$.
Performance is evaluated using Rouge-L~\citep{lin2004rouge} for Korean-Legal-Summary and accuracy score for another two tasks. For comparison, we adopt SliceGPT~\citep{ashkboosslicegpt} as a baseline, which replaces each weight matrix with a smaller proxy matrix without considering the specific use-case requirements.

\begin{figure}[t]
  \centering
    \includegraphics[width=0.48\textwidth]{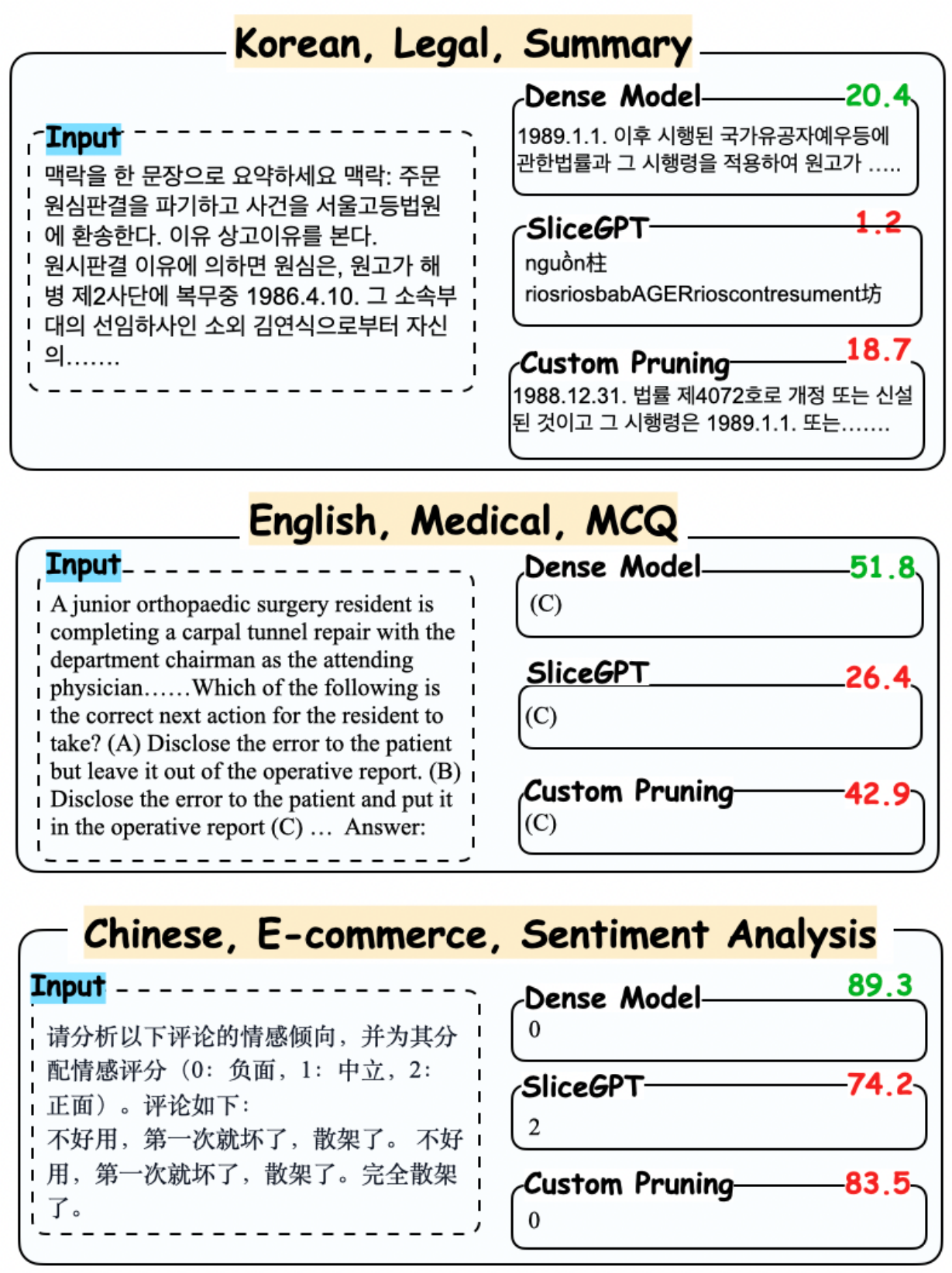}
  \caption{Concrete examples of applying \texttt{Cus-Prun} to prune $25\%$ of Llama3-8B-Base's parameters into three-dimensional expert models. Numbers above each box indicate performance on the \textbf{whole} test set, with the first evaluated by Rouge-L, and the other two by accuracy. }
  \label{fig:korean}
\end{figure}

\begin{table*}[t]
\caption{Main Results of \texttt{Cus-Prun} on multilingual setting with a pruning ratio of $25\%$, where ``general capability'' is tested in English and averaged across several expert models, while  ``specific capability'' is averaged across languages. Results are expressed in Rouge-L in summarization tasks and in accuracy (\%) for other datasets.}
  \centering
\footnotesize
\setlength{\tabcolsep}{0.8pt}
  \scalebox{0.9}{
  \begin{tabular}{c|l|cccc|ccccc|ccccc|cccc}
\toprule
\quad\quad & \multirow{2}{*}{\textbf{\normalsize{Method}}}  & 
  \multicolumn{4}{c}{\textbf{\normalsize{General Capability}}} \vline & \multicolumn{5}{c}{\textbf{\normalsize{Multilingual Expert}}} \vline & \multicolumn{5}{c}{\textbf{\normalsize{Multidomain Expert }}} \vline & \multicolumn{4}{c}{\textbf{\normalsize{Multitask Expert }}} \\ 
  &  & ARC-c & GSM8K & MMLU & Avg. & MGSM & M3 & XQuAD  & Sum & Avg. & MMCQ & FTQA & TSA & AMSA & Avg. & MSum & ASum & AMCF & Avg.  \\ \midrule
\rowcolor{lightgray}\cellcolor{white}\multirow{5}{*}{\rotatebox{90}{\textbf{Llama3-8B}}} & Dense  & 70.7 & 58.3 & 63.1 & 64.1 &41.2 &49.1 & 63.4 &32.9 &46.7 & 51.8 & 23.9 & 67.1 & 95.9 &   59.8  & 76.6 & 16.2 & 78.2 & 57.0 \\
    & LLMPrun.  & 26.3  & 2.5 & 24.2 & 17.7 & 1.1 & 24.0 & 13.6 & 23.2 & 15.5 & 0.0 & 0.0 & \textbf{61.8} & 76.0 & 34.5 & 62.2 & \textbf{21.8} & \textbf{80.0} & \textbf{54.7} \\
    & SliceGPT  & 41.5 & 0.0 & 24.2 & 21.9 & 0.0 & 14.9 & 16.6 & 8.5 & 10.0 & 22.6 & 0.0 & 41.2 & 53.7 & 29.4 & 7.3 & 2.9 & 51.3 & 20.5 \\
    & ShortGPT  & 38.3 & 0.0 & 28.6  & 22.3 & 0.0 & 26.9 & 0.0 & 2.7 & 7.4 &  3.2 & 0.0 & 38.6 & 35.7 & 19.4 & 4.1 & 4.8 & 43.8 & 17.6 \\
      &\cellcolor{lightblue}\texttt{Cus-Prun}  &\cellcolor{lightblue}\textbf{62.4} & \cellcolor{lightblue}\textbf{37.0} & \cellcolor{lightblue}\textbf{54.7} & \cellcolor{lightblue}\textbf{51.4} & \cellcolor{lightblue}\textbf{30.1} & \cellcolor{lightblue}\textbf{41.5} & \cellcolor{lightblue}\textbf{52.6} & \cellcolor{lightblue}\textbf{31.5} & \cellcolor{lightblue}\textbf{38.9} & \cellcolor{lightblue}\textbf{42.9} & \cellcolor{lightblue}\textbf{20.6} & \cellcolor{lightblue}\textbf{61.8} & \cellcolor{lightblue}\textbf{87.6} & \cellcolor{lightblue}\textbf{53.2} & \cellcolor{lightblue}\textbf{68.4} & \cellcolor{lightblue}12.8 & \cellcolor{lightblue}75.5 & \cellcolor{lightblue}52.2  \\ \midrule
\rowcolor{lightgray}\cellcolor{white}\multirow{5}{*}{\rotatebox{90}{\textbf{Mistral-12B}}}  & Dense  & 82.6 & 68.5 & 50.4 & 67.2 & 51.7 & 43.8 & 49.2 & 25.4 & 42.5 & 54.6 & 26.6 & 69.4 & 92.4 & 60.8 & 88.7 & 3.0 & 78.6 & 56.4  \\
    & LLMPrun.  & 22.5 & 2.7 & 30.7  & 18.6 & 2.1 & 27.8 & 19.0 & \textbf{23.2} & 18.0  & 0.0 & 0.0 & 51.0 & 20.9 & 18.0  & 59.3 & 0.5 & 2.8 & 20.9 \\
    & SliceGPT & 49.4 & 1.9 & 32.1 & 27.8 & 0.8 & 25.1 & 17.4 & 7.8 & 12.8 & 24.9 & 9.2 & 34.2 & 54.3 & 30.7 & 27.4 & 1.3 & 36.3 & 21.7  \\
    & ShortGPT & 37.8 & 0.0 & 33.9 & 23.9 & 2.9 & 27.0 & 18.0 & 5.0 & 13.2 & 31.4 & 7.2 & 39.2 & 52.5 & 32.6 & 26.2 & 0.2 & 42.7 & 23.0  \\
    & \cellcolor{lightblue}\texttt{Cus-Prun}  & \cellcolor{lightblue}\textbf{67.5} & \cellcolor{lightblue}\textbf{43.4} & \cellcolor{lightblue}\textbf{43.8} & \cellcolor{lightblue}\textbf{51.6} & \cellcolor{lightblue}\textbf{34.3} & \cellcolor{lightblue}\textbf{39.2} & \cellcolor{lightblue}\textbf{40.7} & \cellcolor{lightblue}23.1 & \cellcolor{lightblue}\textbf{34.3} & \cellcolor{lightblue}\textbf{47.9} & \cellcolor{lightblue}\textbf{25.1} & \cellcolor{lightblue}\textbf{67.3} & \cellcolor{lightblue}\textbf{83.7} & \cellcolor{lightblue}\textbf{56.0}  & \cellcolor{lightblue}\textbf{83.5} & \cellcolor{lightblue}\textbf{3.4} & \cellcolor{lightblue}\textbf{72.8} & \cellcolor{lightblue}\textbf{50.9} \\ \midrule
\rowcolor{lightgray}\cellcolor{white}\multirow{5}{*}{\rotatebox{90}{\textbf{Llama2-13B}}}  & Dense  & 50.3 & 31.4 & 53.4 & 45.1 & 17.5 & 30.4 & 44.1 & 24.9 & 29.2 & 25.2 & 0.0 & 42.7 & 84.1 & 38.0  & 70.0 & 7.4 & 44.3 & 40.6  \\
    & LLMPrun. &  22.4 & 2.1 & 23.6 & 16.0 & 1.1 & 22.8 & 3.8 & 17.7 & 11.3  & 0.0 & 0.0 & 9.7 & 0.0 & 2.4 & 21.6 & 4.8 & 0.0 & 8.8  \\
    & SliceGPT & 45.9 & 2.4 & 48.7 & 32.3 & 2.8 & 25.3 & 23.4 & 9.9 & 15.5 & 18.7 & 0.0 & 28.4 & 67.3 & 28.6  & 24.5 & 4.9 & 32.9 & 20.8 \\
    & ShortGPT & 39.5 & 3.8 & 37.2 & 26.8 & 2.4 & 23.0 & 24.7 & 11.3 & 15.3 & 16.9 & 0.0 & 34.6 & \textbf{69.8} & 30.3 & 23.8 & 5.2 & 39.1 & 22.7   \\
     & \cellcolor{lightblue}\texttt{Cus-Prun} & \cellcolor{lightblue}\textbf{48.3} & \cellcolor{lightblue}\textbf{20.8} & \cellcolor{lightblue}\textbf{50.0} & \cellcolor{lightblue}\textbf{39.7} & \cellcolor{lightblue}\textbf{12.7} & \cellcolor{lightblue}\textbf{26.2} & \cellcolor{lightblue}\textbf{34.2} & \cellcolor{lightblue}\textbf{24.1} & \cellcolor{lightblue}\textbf{24.3}  & \cellcolor{lightblue}\textbf{25.6} & \cellcolor{lightblue}0.0 & \cellcolor{lightblue}\textbf{38.5} & \cellcolor{lightblue}68.3 & \cellcolor{lightblue}\textbf{33.1} & \cellcolor{lightblue}\textbf{64.5} & \cellcolor{lightblue}\textbf{6.7} & \cellcolor{lightblue}\textbf{42.9} & \cellcolor{lightblue}\textbf{38.0} \\ \midrule
\rowcolor{lightgray}\cellcolor{white}\multirow{5}{*}{\rotatebox{90}{\textbf{Llama3-70B}}} & Dense & 84.1 & 82.7  & 78.8 & 81.9 & 69.5 & 71.1 & 69.1 & 36.6 & 61.6 & 72.1 & 55.3 & 83.6 & 96.2 & 76.8 & 84.2 & 17.3 & 81.8 & 61.1 \\
    & LLMPrun. &  69.1 & 26.0 & 53.2 & 49.4 & 16.8 & 43.7 & 43.0 & 29.0 & 33.1 & 27.3 & 1.0 & 51.0 & 50.3 & 32.4 & 10.2 & 13.7 & 20.6  & 14.8 \\
    & SliceGPT &  65.7 & 0.0 & 54.2 & 40.0 & 3.7 & 44.8 & 33.0 & 21.2 & 25.7 & 57.6 & 27.6 & 68.1 & 59.4 & 53.2 & 58.0 & 14.2 & 68.3 & 46.8  \\
    & ShortGPT &  59.4 & 5.6 & \textbf{75.5}  & 46.8 & 11.9 & 43.1 & 38.8 & 24.0 & 29.5 & 58.4 & 32.2 & 67.5 & 64.9 & 55.8 & 59.6 & 13.9 & 65.8 & 46.4  \\
    & \cellcolor{lightblue}\texttt{Cus-Prun}  & \cellcolor{lightblue}\textbf{68.4} & \cellcolor{lightblue}\textbf{53.2} & \cellcolor{lightblue}66.6 & \cellcolor{lightblue}\textbf{62.7} & \cellcolor{lightblue}\textbf{43.1} & \cellcolor{lightblue}\textbf{57.7} & \cellcolor{lightblue}\textbf{59.8} & \cellcolor{lightblue}\textbf{34.3} & \cellcolor{lightblue}\textbf{48.7} & \cellcolor{lightblue}\textbf{68.2} & \cellcolor{lightblue}\textbf{43.9} & \cellcolor{lightblue}\textbf{81.4} & \cellcolor{lightblue}\textbf{87.8} & \cellcolor{lightblue}\textbf{70.3} & \cellcolor{lightblue}\textbf{80.4} & \cellcolor{lightblue}\textbf{15.7} & \cellcolor{lightblue}\textbf{77.5} & \cellcolor{lightblue}\textbf{57.9} \\ 
\bottomrule
\end{tabular}}
  \label{table:cross_ling}
\vspace{-0.2cm}
\end{table*}

\paragraph{Main Results}

Figure \ref{fig:korean} presents the results and one concrete example for the original dense model, pruned model with SliceGPT, and pruned model with our proposed \texttt{Cus-Prun} method across the three scenarios.
We observe that \texttt{Cus-Prun} largely preserves the performance of the dense model, retraining 92\%, 83\%, and 94\% of the original dense model performance on these three cases respectively.
In contrast, the baseline method SliceGPT, which does not consider specific use cases, largely underperforms compared to \texttt{Cus-Prun}.
Overall, the results demonstrate that our proposed \texttt{Cus-Prun} method could effectively obtain expert models tailored to specific use cases across different languages, domains, and tasks that maintain high performance despite substantial pruning.

\section{Foundational Custom Pruning Assessment}

As demonstrated by preliminary evaluation in Section \ref{sec:pre_eva}, \texttt{Cus-Prun} enables the creation of expert models tailored to specific languages, domains, and tasks. However, existing benchmark datasets may not always align with such specialized requirements for a systematic evaluation. To simplify our evaluation without losing generality, we use two distinct corpora: one focusing independently on a single dimension and another encompassing the remaining two dimensions. This approach allows us to evaluate \texttt{Cus-Prun}'s performance in \textit{multilingual}, \textit{multidomain}, and \textit{multitask} settings. 

Formally, in the multilingual setting, instead of constructing $C_{L_{\text{Exp}}}$, $C_{D_{\text{Exp}}}$ and $C_{T_{\text{Exp}}}$ independently, we can construct two corpora, $C_{L_{\text{Exp}}}$ and $C_{(D_, T)_{\text{Exp}}}$, where $C_{L_{\text{Exp}}}$ helps to identify irrelevant neurons in a specific language ($\tilde{\mathcal{N}}_{L_{\text{Exp}}}$) and $C_{(D_, T)_{\text{Exp}}}$ helps to identify irrelevant neurons in a specific domain-task combination ($\tilde{\mathcal{N}}_{D_{\text{Exp}}\cap T_{\text{Exp}}}$). Formally speaking, \texttt{Cus-Prun} in Equation \ref{equ:cus} is transferred to 
\begin{equation}
\scalebox{0.85}{$ 
\begin{aligned}
\mathcal{LLM}_{\text{Exp}} & = 
\mathcal{LLM} \ominus 
\left\{
\tilde{\mathcal{N}}_{L_{\text{Exp}}} \cap 
\left( 
\tilde{\mathcal{N}}_{D_{\text{Exp}}} \cap \tilde{\mathcal{N}}_{T_{\text{Exp}}}
\right)
\right\} \\
& \equiv 
\mathcal{LLM} \ominus 
\left\{
\tilde{\mathcal{N}}_{L_{\text{Exp}}} \cap 
\tilde{\mathcal{N}}_{D_{\text{Exp}} \cap T_{\text{Exp}}}
\right\}.
\end{aligned}
$}
\label{equ:cross-lingual}
\end{equation}
Note that this simplification is also applicable to $C_{D_{\text{Exp}}}$, $C_{(L, T)_{\text{Exp}}}$ and $C_{T_{\text{Exp}}}$, $C_{(L, D)_{\text{Exp}}}$. 

\subsection{Experiment Setup} 

\paragraph{Benchmarks} Although the primary goal of \texttt{Cus-Prun} is to create expert LLMs tuned to specific use cases, we also evaluate the pruned models on standard general capabilities to ensure minimal performance loss. Specifically, we employ ARC-Challenge~\citep{clark2018think} (5-shots), GSM8K~\citep{cobbe2021training} (5-shots with CoT prompting~\citep{wei2022chain}), and MMLU~\citep{hendrycksmeasuring} (5-shots) to represent models general capability. 
Note that we utilize a generation task and implement CoT prompting method, a more challenging setting that has not been previously evaluated by existing pruning techniques~\citep{songsleb, sharmatruth, yang2024laco,zhang2024finercut}.

\paragraph{Baselines} We employ several state-of-the-art pruning methods as the baseline that do not require post-training after pruning the model. (i) Dense represents the original model without pruning; (ii) LLM-Pruner~\citep{ma2023llm} adopts structural pruning that selectively removes non-critical coupled structures based on gradient information;\footnote{To ensure a fair comparison, we evaluate its performance before post-training, following \citet{men2024shortgpt}.} (iii) SliceGPT~\citep{ashkboosslicegpt} replaces each weight matrix with a smaller dense matrix, reducing the embedding dimension of the network; (iv) ShortGPT~\citep{men2024shortgpt} directly deletes the redundant layers in LLMs
based on an importance score.
We follow the typical pruning setting from these previous studies to set the pruning ratio to $25\%$ for all methods and all models. 

\paragraph{Backbone Models} We choose 4 models from different model series and different sizes, including Llama3-8B-Base~\citep{dubey2024llama}, Mistral-Nemo-Base-2407\footnote{\url{https://huggingface.co/mistralai/Mistral-Nemo-Base-2407}}(short as Mistral-12B), Llama2-13B-Base~\citep{touvron2023llama}, Llama3-70B-Base~\citep{dubey2024llama}.

\subsection{Multilingual Setting}

\paragraph{Dataset}
We employ multiple representative multilingual datasets for multilingual setting, which covers reasoning (MGSM~\citep{shilanguage}, 5-shots), multilingual knowledge (M3Exam~\citep{zhang2023m3exam}, 3-shots, abbreviated as M3), understanding (XQuAD~\citep{artetxe2020cross}, 5-shots), and generation (XLSum~\citep{hasan2021xl}, zero-shots, abbreviated as Sum). Furthermore, we consider three languages spanning a range from high-resource to low-resource including German (De), Chinese (Zh) and Thai (Th). 
More detailed experiment settings are explained in Appendix \ref{sec:appen_setting_1}.

\paragraph{Main Results}

Table \ref{table:cross_ling} summarizes the performance of \texttt{Cus-Prun} on multilingual datasets, with average scores across languages. Detailed breakdown results for each language are shown in Table~\ref{table:cl_ger}, Table~\ref{table:cl_chi} and Table~\ref{table:cl_tha} in Appendix~\ref{sec:appen_cl}.
We can observe that \texttt{Cus-Prun} consistently outperforms baseline pruning methods in achieving expert capabilities while preserving general performance. For example, on Llama3-8B, \texttt{Cus-Prun} achieves a score of \(38.9\) compared to at most \(15.5\) for other methods. Similar improvements are observed for Mistral-12B (\(34.3\) vs. \( 18.0\)), Llama2-13B (\(24.3\) vs. \( 15.5\)), and Llama3-70B (\(48.7\) vs. \( 33.1\)). We can also see that the generation tasks are especially challenging; for instance, \texttt{Cus-Prun} achieves a score of \(30.1\) on MGSM for Llama3-8B, while other methods nearly lose the ability to generate coherent reasoning outputs (often approaching \(0\) accuracy for all but the largest model). Additionally, \texttt{Cus-Prun} performs robustly across both high-resource and low-resource languages. Overall, \texttt{Cus-Prun} excels in both performance and robustness across diverse tasks and languages.

\subsection{MultilDomain Setting}

\paragraph{Dataset} For the multidomain setting, we employ several domain-specific datasets, including medical domain multiply choices questions (MedMCQ~\citep{pmlr-v174-pal22a}, 3-shots, abbreviated as MMCQ), finance domain table question-answering (FinTQA~\citep{chen2021finqa}, 8-shots, abbreviated as FTQA), social media domain sentiment analysis (TSA~\citep{kharde975sentiment}, 3-shots), and e-commerce domain sentiment analysis (AMSA~\citep{zhang2015character}, 3-shots). Moreover, in multidomain setting, our focus is exclusively on the English language. Detailed experiment settings are explained in Appendix \ref{sec:appen_setting_2}.

\paragraph{Main Results} Table \ref{table:cross_ling} shows the performance of \texttt{Cus-Prun} on multidomain setting. We find that \texttt{Cus-Prun} consistently outperforms other pruning methods in both expert and general capabilities. For expert capabilities, \texttt{Cus-Prun} achieves a score of $53.2$ on Llama3-8B, while other pruning methods achieve at most $34.5$. Similar improvements are observed for Mistral-12B (\(56.0\) vs. \(32.6\)), Llama2-13B (\(33.1\) vs. \(30.3\)), and Llama3-70B (\(70.3\) vs. \(55.8\)).

\begin{figure*}[t]
    \centering
    \begin{minipage}{0.4\textwidth}
        \centering
        \captionof{table}{Performance of Chinese-Medical expert model on MCQ task}
        \label{tab:cmexam}
        \scalebox{0.96}{
        \begin{tabular}{l|c|c}
          \hline
          \textbf{Method} & \textbf{General} & \textbf{CMExam} \\\midrule
          \rowcolor{lightgray}\cellcolor{white} Dense & 59.3 & 50.6 \\
          LLM-Pruner & 18.6 & 25.0  \\
          SliceGPT & 27.8 & 26.9 \\
          ShortGPT & 23.9 & 23.7 \\
          \rowcolor{lightblue}\cellcolor{white} \texttt{Cus-Prun} & \textbf{52.4} & \textbf{48.7} \\
          \hline
        \end{tabular}}
    \end{minipage}
    \hfill
    \begin{minipage}{0.58\textwidth}
        \centering
        \includegraphics[width=\textwidth]{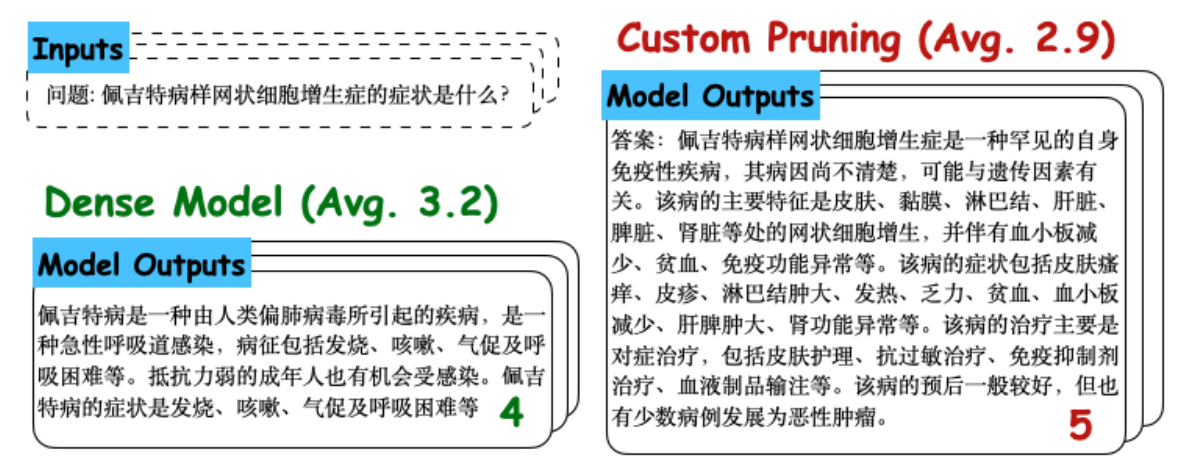}
        \caption{Chinese Medical LLM performance. Numbers are quality on the \textbf{whole} test set evaluated by GPT4.}
        \label{fig:huatuo}
    \end{minipage}
\end{figure*}

\subsection{MultiTask Setting}

\paragraph{Dataset} For the multitask setting, we employ several task-specific datasets, including the medical summarization task (MedSum~\citep{abacha2019summarization}, 3-shots, abbreviated as MSum), summarization task in e-commerce (Amazon Summary~\citep{wang2022supernaturalinstructionsgeneralizationdeclarativeinstructions, brüelgabrielsson2024compressserveservingthousands}, 3-shots, abbreviated as ASum), counterfactual task in e-commerce (Amazon Counterfactual~\citep{o2021wish}, 3-shots, abbreviated as AMCF). Similarly, in multitask setting scenarios, our focus is exclusively on the English language. Detailed experiment settings are explained in Appendix \ref{sec:appen_setting_3}.

\paragraph{Main Results} Table \ref{table:cross_ling} shows the performance of \texttt{Cus-Prun} on multitask setting. We find that except for LLM-Pruner under Llama3-8B, \texttt{Cus-Prun} outperforms other pruning methods in both expert and general capabilities. For expert tasks, \texttt{Cus-Prun} achieves a score of \(50.9\) on \texttt{Mistral-12B}, significantly outperforming the highest score of \(23.0\) from other methods. Likewise, it attains \(38.0\) on \texttt{Llama2-13B} and \(57.9\) on \texttt{Llama3-70B}, both markedly higher than the corresponding maximum scores of \(22.7\) and \(46.8\) achieved by competing methods.

\subsection{Analysis on Aggressive Pruning Ratio}

To optimize for specialized tasks rather than maintaining general capabilities, we employ more aggressive pruning ratios. We combine layer pruning with our custom neuron pruning method in Algorithm \ref{algo:adaptive_pruning} and evaluate the approach on M3Exam, MedMCQ, and Amazon Counterfactual (AMContFact) datasets using Llama3-8B. Detailed results are shown in Table \ref{table:prune_more}. We find that \texttt{Cus-Prun} consistently maintains the model's capabilities even at higher pruning ratios. Specifically, when the pruning ratio is increased to $45\%$, ShortGPT nearly loses the capability of generating meaningful answers, while \texttt{Cus-Prun} still achieves scores of $48.4$ on MMLU and $50.6$ on expert capabilities.

\begin{table}[ht]
\caption{Aggressive pruning ratio on Llama3-8B.}
\centering
\setlength{\tabcolsep}{3pt}
 \scalebox{0.9}{
\begin{tabular}{l|c|c|c|c}
      \hline
      \textbf{Method} & \textbf{Ratio}  & \textbf{Speedup} & \textbf{MMLU} & \textbf{Expert} \\\midrule
     \rowcolor{lightgray}\cellcolor{white}Dense & 0.0 & 1$\times$ & 63.1 & 59.7 \\\midrule
      ShortGPT & 25.0 & 1.3$\times$ & 28.6 & 24.6 \\
     \rowcolor{lightblue}\cellcolor{white}\texttt{Cus-Prun} & 25.0 & 1.3$\times$ & \textbf{51.9} & \textbf{53.3} \\\midrule
      ShortGPT & 34.2 & 1.5$\times$ & 20.8 & 18.5 \\
     \rowcolor{lightblue}\cellcolor{white}\texttt{Cus-Prun} & 35.0 & 1.5$\times$ & \textbf{50.2} & \textbf{51.4} \\\midrule
      ShortGPT & 43.8 & 1.8$\times$ & 7.9 & 10.2 \\
     \rowcolor{lightblue}\cellcolor{white}\texttt{Cus-Prun} & 45.0 & 1.8$\times$ & \textbf{48.4} & \textbf{50.6} \\
      \hline
    \end{tabular}}
    \label{table:prune_more}
\end{table}

\section{Adaptive Custom Pruning Assessment}

In this section, we evaluate the generality of \texttt{Cus-Prun} in dynamic scenarios, including specific expert models in two and one dimensions, as described in Section \ref{sec:adaptive}.

\subsection{Two Dimensions Specific Expert Model}

\paragraph{Experiment Settings} We use the Chinese-Medical setting as a concrete example of a two-dimensional expert model designed to perform a wide range of medical tasks in Chinese. 
We adopt Mistral-12b as the backbone model and utilize corpus from Wikipedia for Chinese content and general medical corpus for medical knowledge. The performance of the target Chinese-Medical expert model is evaluated on two datasets: CMExam~\citep{liu2023benchmarking} (5-shots), a Chinese medical multiple-choice question dataset, and HuatuoQA~\citep{li2023huatuo26m}, a Chinese medical question-answering dataset. We assess the performance on CMExam using accuracy metrics. For the latter, we sample a sub-testset of size $100$ and use GPT-4 as the evaluator, which assigns a score from $0$ to $5$, representing its quality from low to high. Detailed prompts are listed in Appendix \ref{sec:appen_prompt}.

\paragraph{Main Results} Table \ref{tab:cmexam} presents the performance of the Chinese-Medical LLM on CMExam and its general capabilities. Our results indicate that the expert model pruned using \texttt{Cus-Prun} outperforms models obtained through other pruning methods. Specifically, \texttt{Cus-Prun} achieves a score of $48.7$ on CMExam, while its general capability score is $52.4$. These results compare favorably to the dense model, which scores $50.6$ on CMExam and $59.3$ on general capabilities. On the contrary, other pruning methods nearly lose the general and specific capabilities. Furthermore, Figure \ref{fig:huatuo} shows a concrete example of Chinese-Medical LLM performance on medical question-answering. We find that \texttt{Cus-Prun} can produce smaller expert models that maintain their expert capabilities, as demonstrated by its performance score of $2.9/5.0$ compared to $3.2/5.0$ for the dense model.

\subsection{One Dimension Specific Expert Model}

\begin{figure*}[t]
  \centering
    \includegraphics[width=1.0\textwidth]{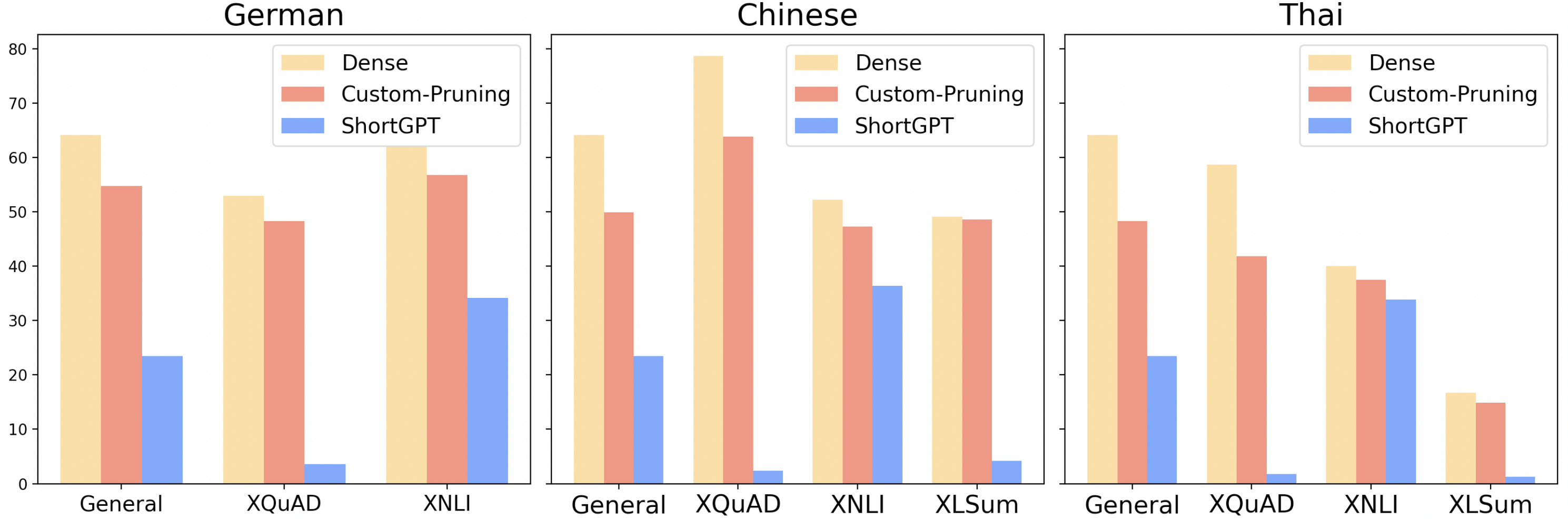}
  \caption{Performance of \texttt{Cus-Prun} in obtaining language-specific models.}
  \label{fig:lang-spec}
\end{figure*}

\paragraph{Experiment Settings}

For evaluating the pruning method under a one-dimensional expert model setting, we focus on language-specific pruning, showing how to transform a dense model into language-specific variants.
We consider three linguistically diverse languages: German, Chinese, and Thai. We conduct experiments based on the Llama3-8b model.
To identify language-specific (while domain- and task-agnostic) neurons, we employ a diverse range of corpora, including Wikipedia, MGSM, and M3Exam, ensuring coverage of various domains and tasks. The effectiveness of our pruning technique is then evaluated using three held-out multilingual datasets including XQuAD~\citep{artetxe2020cross}, XNLI~\citep{conneau2018xnli}, and XSum ~\citep{narayan2018don}.

\paragraph{Main Results} 
Figure \ref{fig:lang-spec} illustrates the performance of language-specific models using \texttt{Cus-Prun}. By pruning $25\%$ of the neurons from the original model, \texttt{Cus-Prun} not only retains general performance but also preserves language-specific capabilities. For instance, the German-specific model scores $54.7$ in general capabilities, $48.3$ on XQuAD, and $56.8$ on XNLI, compared to the dense model's scores of $64.1$, $52.9$, and $62.0$, respectively. This trend is consistent for Chinese and Thai models as well. In contrast, ShortGPT struggles to maintain the model's capabilities, particularly in XQuAD and XLSUm, which require generative abilities.

\section{Related Work}

\paragraph{LLM Compression} Given the high costs of training, inferencing, and tuning LLMs, many studies focus on model compression methods, including compression~\citep{zhu2023survey}, quantization~\citep{xu2023qa, dettmers2024qlora, lin2024awq, li2024evaluating}, and pruning~\citep{wang2019structured}. Sparsity-based structural pruning enhances GPU efficiency with sparse structures but doesn't always reduce parameter counts \citep{li2022parameter, li2023losparse, kurz2024language, zhao2024convex, huang2024pruning}. Unstructured pruning reduces parameters while maintaining performance, using post-training techniques \citep{ma2023llm, xiasheared, muralidharan2024compact} or coarse methods like parameter approximation \citep{zhao2024convex}, layer removal \citep{men2024shortgpt}, or structural elimination \citep{zhang2024finercut}. However, these often fail to preserve domain- or task-specific capabilities, limiting their utility for specialized scenarios.

\paragraph{Customizing Model} Customizing LLMs is essential for addressing language-specific challenges, domain-specific needs, and task-specific applications \citep{cui2023efficient, yang2024laco, li2023huatuo26m, roziere2023code, li2023large, azerbayevllemma, alves2024tower,naacl-sentiment}. However, customization often demands extensive fine-tuning with curated data, making it resource-intensive. This highlights the need for efficient methods to quickly create robust, expert models tailored to diverse industries without compromising quality.

\section{Conclusion}

LLMs deliver impressive capabilities but incur high computational costs. Efficient pruning of redundant parameters is vital for conserving resources and improving inference speed, especially for specialized models. Our method, \texttt{Cus-Prun}, generates smaller expert models without post-training by pruning irrelevant neurons across ``language'', ``domain'', and ``task'' dimensions. This finer-grained approach outperforms existing techniques on three-dimensional models and can adapt to realistic scenarios, such as language-domain or language-specific models.

\section*{Limitation}

Despite the promising results of Cus-Prun, several limitations should be noted. First, while our method leverages three dimensions (language, domain, and task) for pruning, certain crucial restrictions cannot be fully captured within this framework, such as variations in query format or input structure. Second, whether pruned base models can effectively undergo post-training remains an open question that requires further investigation. This uncertainty about post-training capabilities could limit the model's adaptability to new scenarios or requirements after pruning. These limitations suggest important directions for future research, including exploring additional dimensions for more comprehensive pruning strategies and investigating the relationship between pruning and post-training effectiveness.

\section*{Acknowledgement}
This research is supported by the Ministry of Education, Singapore, under its Academic Research Fund (AcRF) Tier 1 grant, and funded through the SUTD Assistant Professorship Scheme (SAP 2025\_001).
This material is based upon work partially supported by the Air Force Office of Scientific Research under award number FA2386-24-1-4011, and this research is partially supported by the Singapore Ministry of Education Academic Research Fund Tier 1 (Award No: T1 251RES2207).

\bibliography{custom}

\appendix
\section{Appendix}

\subsection{GPT-4o Prompts}\label{sec:appen_prompt}
The prompts used for generating documents and evaluating answer quality are presented in Table~\ref{table:testset}.

\begin{table}[ht]
  \centering
\footnotesize
  \scalebox{1}{
  \begin{tabular}{lp{5.3cm}}
    \toprule
   \textbf{Task} & \textbf{Prompt} \\\midrule
   Generation & Generate a text document in \{language\}/\{domain\}/\{task\}. Make sure the documents is not fixed to one \{language\}/\{domain\}/\{task\} or \{language\}/\{domain\}/\{task\}. Ensure the content is clear, concise, and appropriate for the specified request. Use professional and domain-specific terminology where necessary. \\\midrule
   Evaluation & Evaluate the quality of the given answer to the question. Provide a score from 0 to 5, where 0 represents very low quality and 5 represents very high quality. Question: \{question\} Answer: \{answer\}.\\
    \bottomrule
  \end{tabular}}
\caption{GPT-4o prompts for generating documents and evaluating answer quality.}  \label{table:testset}
\end{table}

\subsection{Detailed Results for Multilingual}\label{sec:appen_cl}

Detailed results for multilingual settings can be found in Table \ref{table:cl_ger}, Table \ref{table:cl_chi} and Table \ref{table:cl_tha} for German, Chinese and Thai, respectively.

\begin{table*}[hbpt]
\caption{Main Results of \texttt{Cus-Prun} on Germany with a pruning ratio of $25\%$, where ``general capability'' is tested in English and averaged across several expert models, while  ``specific capability'' is averaged across languages. Results are expressed in Rouge-L in XLSum and in accuracy (\%) for other datasets.}
  \centering
\footnotesize
\setlength{\tabcolsep}{3pt}
  \scalebox{1.0}{
  \begin{tabular}{l|l|cccc|ccccc}
\toprule
\multirow{2}{*}{\textbf{\normalsize{Model}}} & \multirow{2}{*}{\textbf{\normalsize{Method}}} & 
  \multicolumn{4}{c}{\textbf{\normalsize{General Capability}}} \vline & \multicolumn{5}{c}{\textbf{\normalsize{Expert Capability}}} \\ 
  &  & ARC-c & GSM8K & MMLU & Avg. & MGSM & M3Exam & XQuAD  & XLSum & Avg.  \\ \midrule
\multirow{4}{*}{Llama3-8B} &
    Dense  & 70.7 & 58.3 & 63.1 & 64.1 & 44.8  & - & 52.9 & - & 48.8   \\
    & LLMPrun.   & 26.3  & 2.5 & 24.2 & 17.7 & 0.0 & - & 11.0 & - & 5.5 \\
    & SliceGPT & 41.5 & 0.0 & 24.2 & 21.9 & 0.0 & - & 9.8 & - & 4.9  \\
    & ShortGPT   & 38.3 & 0.0 & 28.6  & 22.3 & 0.0 & - & 0.0 & - & 0.0  \\
    & \texttt{Cus-Prun}    & 61.4 & 38.9 & 54.5 & \textbf{51.6} & 32.8 & - & 49.6 & - & \textbf{41.2}  \\ \midrule
\multirow{4}{*}{Mistral-12B}  & Dense   & 82.6 & 68.5 & 50.4 & 59.3 & 56.8  & - & 41.2  & - & 49.0   \\
    & LLMPrun.  & 22.5 & 2.7 & 30.7  & 18.6 & 2.4  & - & 13.4 & - & 7.9  \\
    & SliceGPT & 49.4 & 1.9 & 32.1 & 27.8 & 0.8 & - & 15.5 & - & 8.2 \\
    & ShortGPT & 37.8 & 0.0 & 33.9 & 23.9 & 3.6 & - & 20.3 & - & 12.0  \\
    & \texttt{Cus-Prun}  & 64.6 & 39.7 & 43.2 & \textbf{49.2} & 31.6  & - & 35.9 & - & \textbf{33.8}  \\ \midrule
\multirow{4}{*}{Llama2-13B}  & Dense  & 50.3 & 31.4 & 53.4 & 45.1 & 24.4 & - & 40.3  & - & 32.3  \\
    & LLMPrun.  & 22.4 & 2.1 & 23.6 & 16.0  & 2.0 & - & 5.7 & - & 3.9 \\
    & SliceGPT & 45.9 & 2.4 & 48.7 & 32.3 & 3.6 & - & 18.1 & - & 10.9 \\
    & ShortGPT & 39.5 & 3.8 & 37.2 & 26.8 & 2.8 & - & 27.2 & - & 15.0  \\
    & \texttt{Cus-Prun}  & 47.6 & 19.8 & 49.9 & \textbf{39.1} & 18.4 &  - & 31.7 &  - & \textbf{25.0} \\ \midrule
\multirow{4}{*}{Llama3-70B} & Dense & 84.1 & 82.7  & 78.8 & 81.9 & 74.8 & - & 58.2 & - & 66.5  \\
    & LLMPrun. &  69.1 & 26.0 & 53.2 & 49.4 & 18.0 & - & 27.3 & - & 22.7 \\
    & SliceGPT &  65.7 & 0.0 & 54.2 & 40.0 &  0.0 & - & 17.3 & - & 8.7    \\
    & ShortGPT &  59.4 & 5.6 & 75.5  & 46.8 & 9.6 & - & 31.5 & - & 20.6   \\
    & \texttt{Cus-Prun}  & 66.8 & 59.3 & 69.1 & \textbf{65.1} & 48.2   & - & 53.9 & - & \textbf{51.1}  \\ 
\bottomrule
\end{tabular}}
  \label{table:cl_ger}
\end{table*}

\begin{table*}[hbpt]
\caption{Main Results of \texttt{Cus-Prun} on Chinese with a pruning ratio of $25\%$, where ``general capability'' is tested in English and averaged across several expert models, while  ``specific capability'' is averaged across languages. Results are expressed in Rouge-L in XLSum and in accuracy (\%) for other datasets.}
  \centering
\footnotesize
\setlength{\tabcolsep}{3pt}
  \scalebox{1.0}{
  \begin{tabular}{l|l|cccc|ccccc}
\toprule
\multirow{2}{*}{\textbf{\normalsize{Model}}} & \multirow{2}{*}{\textbf{\normalsize{Method}}} & 
  \multicolumn{4}{c}{\textbf{\normalsize{General Capability}}} \vline & \multicolumn{5}{c}{\textbf{\normalsize{Specific Capability}}} \\ 
  &  & ARC-c & GSM8K & MMLU & Avg. & MGSM & M3Exam & XQuAD  & XLSum & Avg.  \\ \midrule
\multirow{4}{*}{Llama3-8B} &
    Dense & 70.7 & 58.3 & 63.1 & 64.1 &  43.6 & 55.1 & 78.7  & 49.1 & 56.6  \\
    & LLMPrun.   & 26.3  & 2.5 & 24.2 & 17.7 & 2.4 & 23.6   & 21.3 & 32.8 & 20.0 \\
    & SliceGPT  & 41.5 & 0.0 & 24.2 & 21.9 & 0.0 & 17.4 & 23.5 & 8.3 & 12.3  \\
    & ShortGPT   & 38.3 & 0.0 & 28.6  & 22.3 & 0.0 & 28.3 & 0.0 & 3.1 & 7.9  \\
    & \texttt{Cus-Prun}   & 60.5 & 25.7 & 49.4 & \textbf{45.2} & 36.0 & 44.7 & 65.6 & 46.3 & \textbf{48.2}  \\ \midrule
\multirow{4}{*}{Mistral-12B}  & Dense  & 82.6 & 68.5 & 50.4 & 59.3 &  53.2 & 47.8  & 62.2 & 33.0 & 49.1 \\
    & LLMPrun.  & 22.5 & 2.7 & 30.7  & 18.6 & 2.8 & 30.7   & 31.8 & 32.6 & 24.5 \\
    & SliceGPT  & 49.4 & 1.9 & 32.1 & 27.8 & 1.6 & 26.4 & 28.3 & 10.8 & 16.8  \\
    & ShortGPT  & 37.8 & 0.0 & 33.9 & 23.9 & 4.4 & 28.2 & 29.1 & 7.2 & 17.2  \\
    & \texttt{Cus-Prun}  & 68.3 & 43.2 & 39.5 & \textbf{50.3} & 38.4 & 40.7 & 50.6 & 30.3 & \textbf{40.0}  \\ \midrule
\multirow{4}{*}{Llama2-13B}  & Dense  & 50.3 & 31.4 & 53.4 & 45.1 & 21.6 & 36.5 & 59.8 & 35.3 & 38.3 \\
    & LLMPrun.  & 22.4 & 2.1 & 23.6 & 16.0  & 1.2 & 23.3 & 3.8 & 25.1 & 13.4\\
    & SliceGPT  & 45.9 & 2.4 & 48.7 & 32.3 & 4.8 & 24.5 & 28.4 & 11.2 & 17.2 \\
    & ShortGPT  & 39.5 & 3.8 & 37.2 & 26.8 & 4.4 & 22.9 & 24.6 & 13.7 & 16.4  \\
    & \texttt{Cus-Prun} & 48.6 & 20.7 & 51.9 & \textbf{40.4} & 14.8 & 28.2 & 47.3 & 34.4 & \textbf{31.2}  \\ \midrule
\multirow{4}{*}{Llama3-70B} & Dense   & 84.1 & 82.7 & 78.8 & 81.9 & 68.4 & 76.1 & 81.3 & 55.3 & 70.3  \\
    & LLMPrun. &  69.1 & 26.0 & 53.2 & 49.4  & 16.8 & 47.5 & 56.1 & 41.3 & 40.4  \\
    & SliceGPT & 65.7 & 0.0 & 54.2 & 40.0 & 6.4 & 48.3 & 42.2 & 29.3 & 31.6    \\
    & ShortGPT & 59.4 & 5.6 & 75.5 & 46.8 & 12.4 & 45.5 & 44.6 & 36.1 & 34.7  \\
    & \texttt{Cus-Prun}  & 72.3 & 48.5 & 65.2 & \textbf{62.0} & 40.8 & 61.7 & 66.9 & 51.6 & \textbf{55.3} \\ 
\bottomrule
\end{tabular}}
  \label{table:cl_chi}
\end{table*}

\begin{table*}[hbpt]
\caption{Main Results of \texttt{Cus-Prun} on Thai with a pruning ratio of $25\%$, where ``general capability'' is tested in English and averaged across several expert models, while  ``specific capability'' is averaged across languages. Results are expressed in Rouge-L in XLSum and in accuracy (\%) for other datasets.}
  \centering
\footnotesize
\setlength{\tabcolsep}{3pt}
  \scalebox{1.0}{
  \begin{tabular}{l|l|cccc|ccccc}
\toprule
\multirow{2}{*}{\textbf{\normalsize{Model}}} & \multirow{2}{*}{\textbf{\normalsize{Method}}} & 
  \multicolumn{4}{c}{\textbf{\normalsize{General Capability}}} \vline & \multicolumn{5}{c}{\textbf{\normalsize{Specific Capability}}} \\ 
  &  & ARC-c & GSM8K & MMLU & Avg. & MGSM & M3Exam & XQuAD  & XLSum & Avg.  \\ \midrule
\multirow{4}{*}{Llama3-8B} &
    Dense  & 70.7 & 58.3 & 63.1 & 64.1 & 35.2 & 43.0 & 58.7 & 16.7 & 38.4  \\
    & LLMPrun.   & 26.3  & 2.5 & 24.2 & 17.7 & 0.8 & 24.4  & 8.4 & 13.5 & 11.8  \\
    & SliceGPT  & 41.5 & 0.0 & 24.2 & 21.9 & 0.0 & 12.3 & 16.6 & 8.7 & 9.4  \\
    & ShortGPT  & 38.3 & 0.0 & 28.6  & 22.3 & 0.0 & 25.4 & 0.0 & 2.3 & 6.9  \\
    & \texttt{Cus-Prun}  & 58.9 & 31.2 & 52.4 & \textbf{47.5} & 21.6 & 38.3 & 42.6 & 16.8 & \textbf{29.8}  \\ \midrule
\multirow{4}{*}{Mistral-12B}  & Dense  & 82.6 & 68.5 & 50.4 & 59.3 & 45.2 & 39.9 & 44.1  & 17.8 & 36.8  \\
    & LLMPrun. & 22.5 & 2.7 & 30.7  & 18.6 & 1.2 & 24.8  & 11.9 & 13.7 & 12.9 \\
    & SliceGPT & 49.4 & 1.9 & 32.1 & 27.8 & 0.0 & 23.8 & 8.4 & 4.7 & 12.3 \\
    & ShortGPT  & 39.5 & 3.8 & 37.2 & 26.8 & 0.8 & 25.7 & 4.7 & 2.8 & 8.5 \\
    & \texttt{Cus-Prun} & 68.2 & 35.8 & 47.6 & \textbf{50.5} & 32.8 & 37.7 & 35.6 & 15.9 & \textbf{30.5}  \\ \midrule
\multirow{4}{*}{Llama2-13B} & Dense & 50.3 & 31.4 & 53.4 & 45.1 & 6.4 & 24.3 & 28.3 & 14.5 & 18.4 \\
    & LLMPrun.  & 22.4 & 2.1 & 23.6 & 16.0  & 0.0 & 22.3 & 1.8 & 10.2 & 8.6 \\
    & SliceGPT  & 45.9 & 2.4 & 48.7 & 32.3 & 0.0 & 26.2 & 23.7 & 8.6 & 14.6 \\
    & ShortGPT  & 39.5 & 3.8 & 37.2 & 26.8 & 0.0 & 23.1 & 22.3 & 8.9 & 13.6  \\
    & \texttt{Cus-Prun} & 47.8 & 20.9 & 50.7 & \textbf{39.8} & 4.8 & 24.2 & 23.6 & 13.8 & \textbf{16.6} \\ \midrule
\multirow{4}{*}{Llama3-70B} & Dense  & 84.1 & 82.7 & 78.8 & 81.9 & 65.2 & 66.1 & 67.8 & 17.8 & 54.2  \\
    & LLMPrun. &  69.1 & 26.0 & 53.2 & 49.4  & 15.6 & 39.9 & 29.8 & 16.6 & 25.5 \\
    & SliceGPT & 65.7 & 0.0 & 54.2 & 40.0 & 4.8 & 41.3 & 39.6 & 13.2 & 24.7  \\
    & ShortGPT & 59.4 & 5.6 & 75.5 & 46.8 & 13.7 & 40.7 & 40.4 & 11.9 & 26.7   \\
    & \texttt{Cus-Prun}  & 73.3 & 58.7 & 68.4 & \textbf{66.8} & 40.4 & 53.6 & 58.5 & 16.9 & \textbf{42.4}  \\ 
\bottomrule
\end{tabular}}
  \label{table:cl_tha}
\end{table*}

\subsection{Experiments Detailed Settings}\label{sec:appen_setting}

\subsubsection{Multilingual Settings}\label{sec:appen_setting_1}

\paragraph{Experiment Details} For multilingual setting, we can obtain two corpora: $C_{L_{\text{Exp}}}$ and $C_{(D, T)_{\text{Exp}}}$. The first corpus contains samples in a specific language across various domains and tasks, while the second corpus contains samples from a specific domain-task combination in other languages, i.e., the target dataset in other languages. Specifically, for $C_{L_{\text{Exp}}}$ we employ Wikipedia\footnote{\url{https://huggingface.co/datasets/wikimedia/wikipedia}} to construct language-specific corpus covering various domains and tasks. For $C_{(D, T)_{\text{Exp}}}$, we employ the corresponding datasets in English, including GSM8K~\citep{cobbe2021training} for MGSM, the English split of M3Exam\footnote{M3Exam is language-specific and does not utilize a translated parallel corpus.} for M3Exam, SQuAD~\citep{rajpurkar2016squad} for XQuAD, and XSum~\citep{narayan2018don} for XLSum. 

Hyperparameters, including the sizes of $C_{L_{\text{Exp}}}$ and $C_{(D, T)_{\text{Exp}}}$, are determined using the validation set of the XLSum dataset and then applied to testsets in other multilingual datasets. Furthermore, accuracy is the metric used for ARC-c, GSM8K, MMLU, MGSM, M3Exam, and XQuAD, while Rouge-L~\citep{lin2004rouge} is used for XLSum. 

\subsubsection{Multidomain Settings}\label{sec:appen_setting_2}

\paragraph{Settings}  For multidomain setting, we can obtain two corpora: $C_{D_{\text{Exp}}} = \{(L, D_{\text{Exp}}, T)|L\in\mathbb{L}, T\in\mathbb{T}\}$ and $C_{(L, T)_{\text{Exp}}}=\{(D, (L, T)_{\text{Exp}})|D\in\mathbb{D}\}$. The first corpus contains samples in a specific domain across various languages and tasks, while the second corpus contains samples from a specific language-task combination across different domains, i.e., the target dataset in other domains. Specifically, for $C_{D_{\text{Exp}}}$ we employ specific domain corpus, including English split of medical corpus~\citep{garcia2024medical} for medical domain, general finance corpus for finance domain\footnote{\url{https://huggingface.co/datasets/gbharti/finance-alpaca}}, general Twitter corpus~\citep{kharde975sentiment}, and English split of Amazon corpus~\citep{marc_reviews}. For $C_{(L, T)_{\text{Exp}}}$, we employ the corresponding datasets in general domains, including CommonsenseQA~\citep{talmor2019commonsenseqa} for MedMCQ, open table question-answering OTT-QA~\citep{chen2020open} for FinTQA, general sentiment analysis~\citep{attia2018multilingual} for TSA and AMSA.

\paragraph{Experiment Details} Hyperparameters, including the sizes of $C_{D_{\text{Exp}}}$ and $C_{(L, T)_{\text{Exp}}}$, are determined using the validation set of the Amazon sentiment analysis dataset and then applied to testsets in other multidomain datasets. Furthermore, accuracy is the metric used for all datasets.

\subsubsection{Multitask Settings}\label{sec:appen_setting_3}

\paragraph{Settings} For multitask setting, we can obtain two corpora: $C_{T_{\text{Exp}}} = \{(L, D, T_{\text{Exp}})|L\in\mathbb{L}, D\in\mathbb{D}\}$ and $C_{(L, D)_{\text{Exp}}}=\{(T, (L, S)_{\text{Exp}})|T\in\mathbb{T}\}$. The first corpus contains samples in a specific task across various languages and domains, while the second corpus contains samples from a specific language-domain combination across different tasks, i.e., the target dataset in other tasks. Specifically, for $C_{T_{\text{Exp}}}$ we employ specific task corpus, including XSum corpus~\citep{abacha2019summarization} for summarization task, general conterfact corpus\footnote{\url{https://huggingface.co/datasets/azhx/counterfact-easy}} for counterfactual task. For $C_{(L, D)_{\text{Exp}}}$, we employ the corresponding datasets in other tasks, including MedQCQ~\citep{pmlr-v174-pal22a} for MedSum, AMSA~\citep{zhang2015character} for AMSum and AMContFact.

\paragraph{Experiment Details} Hyperparameters, including the sizes of $C_{T_{\text{Exp}}}$ and $C_{(L, D)_{\text{Exp}}}$, are determined using the validation set of the Amazon counterfactual dataset and then applied to testsets in other multitask setting  datasets. Furthermore, accuracy is the metric used for ARC-c, GSM8K, MMLU, and AMContFact, while Rouge-L~\citep{lin2004rouge} is used for MedSum and AMSum.

\end{document}